%% file: main.tex
\documentclass{article}

\PassOptionsToPackage{numbers}{natbib}

\usepackage[main, final]{neurips_2025}

\usepackage[utf8]{inputenc} % allow utf-8 input
\usepackage[T1]{fontenc}    % use 8-bit T1 fonts
\usepackage{url}            % simple URL typesetting
\usepackage{booktabs}       % professional-quality tables
\usepackage{amsfonts}       % blackboard math symbols
\usepackage{graphicx}       % for including graphics
\usepackage{nicefrac}       % compact symbols for 1/2, etc.
\usepackage{microtype}      % microtypography
\usepackage{xcolor}         % colors
\usepackage{amsmath, amssymb, bm}
\usepackage{algorithm}  
\usepackage[noend]{algpseudocode}  
\usepackage{algorithmicx}
\usepackage{array}
\usepackage{colortbl}
\usepackage{hyperref}       % hyperlinks
\usepackage{pifont}%
\newcommand{\cmark}{\ding{51}}%
\newcommand{\xmark}{\ding{55}}%
\usepackage{xspace}

\usepackage{graphics} %
\usepackage{epsfig} %
\usepackage{times} %
\usepackage{array}  
\usepackage{multirow} 
\usepackage{makecell}

\usepackage{colortbl}
\usepackage{framed}
\usepackage{wrapfig}  
\usepackage{titletoc}
\usepackage{color} 
\usepackage{caption, subcaption, overpic, textpos}
\usepackage{titlesec}
\usepackage{comment}
\usepackage{fontawesome}
\usepackage{siunitx}
\usepackage{tcolorbox} %
\tcbuselibrary{listings} %

\usepackage{titlesec}
\definecolor{galaxeaOrange}{HTML}{FF5A00}
\hypersetup{
    colorlinks=true,
    linkcolor=blue,
    citecolor=blue,
    urlcolor=blue
}

\makeatletter
\renewcommand{\@notice}{}
\makeatother

\title{Fast-WAM: Do World Action Models Need \\Test-time Future Imagination?}

\author{%
    Tianyuan Yuan$^{1,2}$,
    Zibin Dong$^{1,2}$,
    Yicheng Liu$^{1,2}$,
    Hang Zhao$^{1,2}$
    \\[1ex]
    $^{1}$IIIS, Tsinghua University \hspace{1em}$^{2}$Galaxea AI\\
  \url{https://yuantianyuan01.github.io/FastWAM/} \\
}

\begin{document}

\maketitle
\begin{abstract}
World Action Models (WAMs) have emerged as a promising alternative to Vision-Language-Action (VLA) models for embodied control because they explicitly model how visual observations may evolve under action. Most existing WAMs follow an imagine-then-execute paradigm, incurring substantial test-time latency from iterative video denoising, yet it remains unclear whether explicit future imagination is actually necessary for strong action performance.

In this paper, we ask whether WAMs need explicit future imagination at test time, or whether their benefit comes primarily from video modeling during training. We disentangle the role of video modeling during training from explicit future generation during inference by proposing \textbf{Fast-WAM}, a WAM architecture that retains video co-training during training but skips future prediction at test time. 
We further instantiate several Fast-WAM variants to enable a controlled comparison of these two factors. Across these variants, we find that Fast-WAM remains competitive with imagine-then-execute variants, while removing video co-training causes a much larger performance drop. Empirically, Fast-WAM achieves competitive results with state-of-the-art methods both on simulation benchmarks (LIBERO and RoboTwin) and real-world tasks, without embodied pretraining. It runs in real time with 190\,ms latency, over 4$\times$ faster than existing imagine-then-execute WAMs. These results suggest that the main value of video prediction in WAMs may lie in improving world representations during training rather than generating future observations at test time.
\end{abstract}

\input{intro}
\input{related}
\input{method}
\input{exp}

\input{conclusion}

\newpage
\bibliographystyle{unsrtnat}
\bibliography{va}

%%%%%%%%%%%%%%%%%%%%%%%%%%%%%%%%%%%%%%%%%%%%%%%%%%%%%%%%%%%%

\appendix
\section{Appendix}

\subsection{RoboTwin Detailed Results}
Here we present the per-task results on RoboTwin evaluation in Table~\ref{tab:robotwin_detail}.
\input{tables/tab_robotwin_detail}

\end{document}

%% file: intro.tex
\section{Introduction}

Building general-purpose embodied agents requires policies that can not only map visual observations to actions, but also reason about how the physical world evolves under interaction. This has motivated growing interest in World Action Models (WAMs), which combine future visual prediction and action modeling in a unified framework. Compared with standard Vision-Language-Action (VLA) models, WAMs are appealing because modeling future observations may help capture physical dynamics and task-relevant temporal structure.

Most existing WAMs follow an imagine-then-execute paradigm: they first generate future observations, then predict actions conditioned on the imagined future. While intuitive, this design incurs substantial test-time latency due to iterative video denoising~\cite{pai2025mimicvideo,liang2025videogenerators,lingbot-va2026, ye2026worldactionmodelszeroshot, bi2025motusunifiedlatentaction}. More fundamentally, it remains unclear whether explicit future imagination is actually necessary for strong action performance. The effectiveness of WAMs may stem from two distinct sources: \emph{(1)} the video prediction objective during training, which may help the model acquire stronger physical priors and action-conditioned representations, and \emph{(2)} explicit future generation during inference, which may provide additional foresight for action prediction. Existing WAM systems typically entangle these two factors, making it difficult to determine which one is actually responsible for the observed gains.

In this paper, we revisit this design choice and ask a simple question: \emph{do WAMs need to imagine future observations at test time, or do they benefit primarily from learning to model them during training?} Our key idea is to decouple the video prediction objective used in WAM training from explicit future generation at inference time. If the main value of world modeling lies in shaping better latent representations during training, then a WAM should be able to retain this benefit without paying the test-time cost of future video synthesis.

\begin{figure*}[tbp]
        \centering
\includegraphics[width=1.0\textwidth]{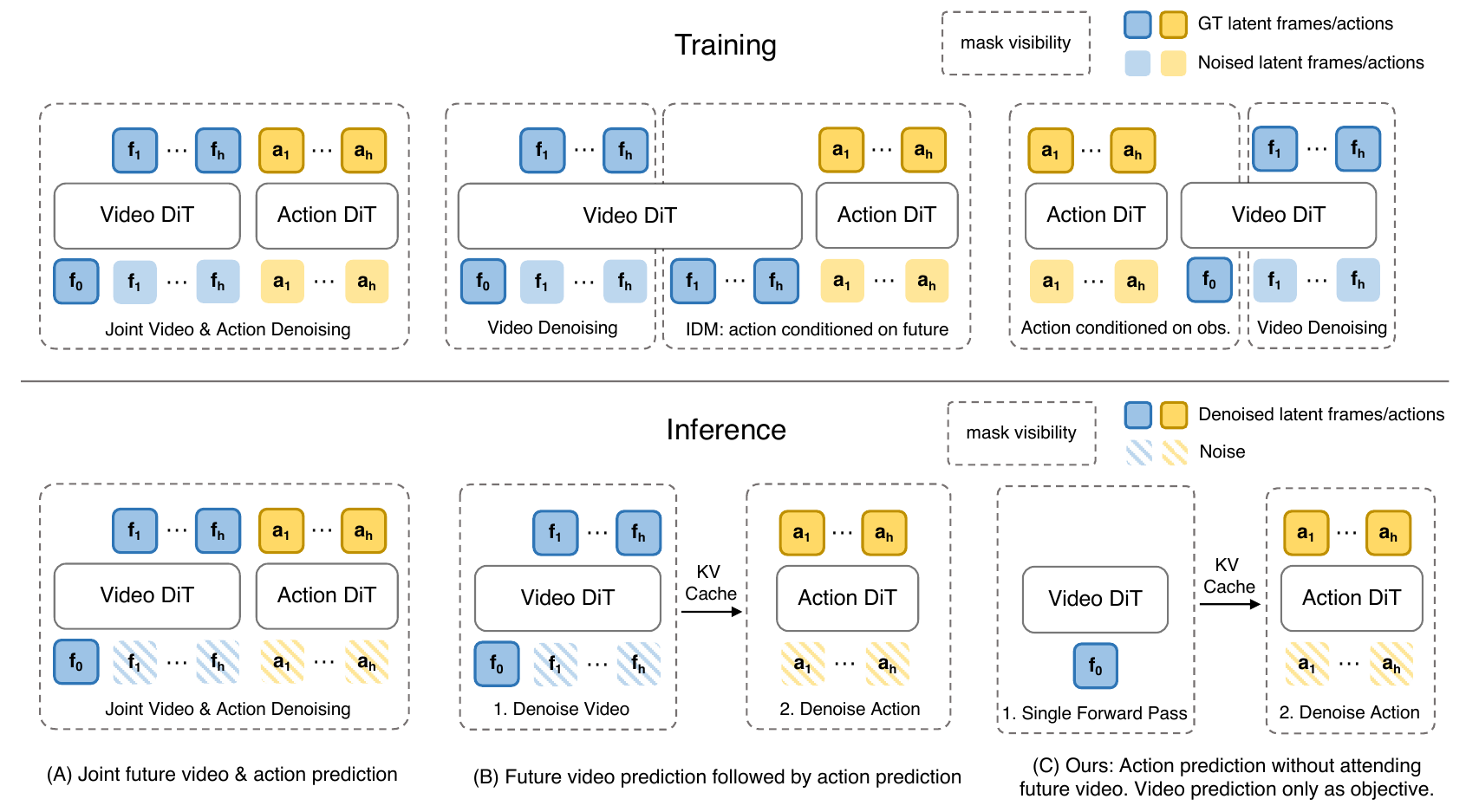} 
    \caption{Three representative WAM paradigms. (A) Joint-modeling WAMs denoise future video and action tokens together. (B) Causal WAMs first generate future observations and then condition action prediction on the generated future representation. (C) Fast-WAM retains video co-training during training but removes explicit future generation at inference time, directly predicting actions from latent world representations in a single forward pass.}
    \label{fig:three_types}
\end{figure*}

Based on this perspective, we propose \textbf{Fast-WAM}, a WAM architecture that preserves video co-training during training but skips future prediction at test time. Instead of using a pretrained video generation model to iteratively synthesize future frames during inference, Fast-WAM repurposes a pretrained video Diffusion Transformer (DiT) as a single-pass world encoder for action generation. Concretely, we build Fast-WAM with a Mixture-of-Transformer (MoT) architecture with shared attention, consisting of a video DiT and an action expert DiT, as illustrated in Figure~\ref{fig:three_types}(C). During training, the video prediction objective shapes the video DiT to encode physically meaningful motion and interaction structure. During inference, the video DiT processes the observation context in a single forward pass and provides latent world representations for action denoising, avoiding explicit future video denoising and enabling efficient real-time control.

To study our central question in a controlled way, we instantiate Fast-WAM into variants that mirror representative imagine-then-execute WAM designs. For simplicity, we focus on single action chunk generation and omit the outer auto-regressive loop. As shown in Figure~\ref{fig:three_types}, existing WAMs can be broadly grouped into two representative paradigms: \emph{(A)} future videos and actions are jointly denoised with shared attention~\cite{ye2026worldactionmodelszeroshot, zhu2025unifiedworldmodelscoupling, bi2025motusunifiedlatentaction}; and \emph{(B)} actions are predicted after, and conditioned on, generated future videos~\cite{lingbot-va2026, feng2025vidarembodiedvideodiffusion, du2023learninguniversalpoliciestextguided}. We also implement a no-video-co-training variant, which serves as a direct control for the role of the training objective itself. Together, these controlled comparisons allow us to isolate the contribution of test-time future imagination from that of video co-training during training.

Experiments on simulation benchmarks (LIBERO and RoboTwin) show that Fast-WAM achieves strong results without any embodied pretraining, demonstrating strong data efficiency. On real-world robotic tasks, Fast-WAM remains highly effective while running at only 190\,ms latency, making it more than 4$\times$ faster than existing imagine-then-execute WAM approaches. More importantly, controlled comparisons show that Fast-WAM stays close to imagine-then-execute variants, while removing the video co-training objective causes a much larger performance drop. These results suggest that the main value of video prediction in WAMs may lie in improving world representations during training rather than in explicitly generating future observations at test time.

Our contributions are three-fold:
\begin{itemize}
    \item We identify and study a basic question in WAMs: whether their gains come primarily from video modeling during training or from explicit future imagination during inference.
    \item We propose Fast-WAM, a WAM architecture that retains video co-training during training while eliminating future prediction at test time, enabling real-time inference.
    \item Through controlled comparisons on simulation and real-world benchmarks, including variants with and without video co-training, we show that much of the benefit of WAMs comes from the video co-training objective itself, while explicit future generation at inference time appears to be less critical than previously assumed.
\end{itemize}

%% file: related.tex
\section{Related Work}

\paragraph{Vision-Language-Action Policies.}
Recent progress in embodied foundation models has been driven by Vision-Language-Action (VLA) policies, which directly map visual observations and language instructions to robot actions using large pretrained vision-language backbones~\cite{kim2024openvla, black2024pi_0, intelligence2025pi_, bjorck2025gr00t, zitkovich2023rt, liu2024rdt, bu2025agibot, shukor2025smolvla, team2025gemini, galaxea2025, wen2025dexvla}. By inheriting strong semantic priors from web-scale pretraining, these models have shown strong generalization across objects, scenes, and language instructions. However, as noted in recent WAM work, standard VLA pretraining is largely based on static image-text data and does not explicitly model how the physical world evolves under action~\cite{lingbot-va2026, ye2026worldactionmodelszeroshot}. Our work is complementary to this line: Fast-WAM preserves a direct-policy interface at test time, similar to VLAs, but learns it under an additional world-modeling objective based on future visual prediction.

\paragraph{World Action Models and Video-based Robot Policies.}
A parallel line of work studies robot control through future visual prediction, using video generation as a way to model environment dynamics and infer actions~\cite{du2023learninguniversalpoliciestextguided, wu2023unleashing, zhou2024robodreamer, feng2025vidarembodiedvideodiffusion, bharadhwaj2024gen2act, won2025dualstreamdiffusionworldmodelaugmented, cheang2024gr2generativevideolanguageactionmodel, jang2025dreamgenunlockinggeneralizationrobot, zhao2025cotvlavisualchainofthoughtreasoning, cen2025rynnvla, cen2025WorldVLA, zhou2025act2goalworldmodelgeneral, zheng2025flarerobotlearningimplicit, dreamvla25}. Recent methods further scale this idea by jointly modeling future video and robot actions in a unified framework~\cite{zhu2025unifiedworldmodelscoupling, liang2025videogenerators, kim2026cosmos, ge2025, pai2025mimicvideo, lingbot-va2026, bi2025motusunifiedlatentaction, ye2026worldactionmodelszeroshot}. We follow~\cite{ye2026worldactionmodelszeroshot} to refer to these models as World Action Models (WAMs), since they leverage world modeling, i.e., predicting future visual states, to support downstream action prediction. Most existing WAMs either follow an imagine-then-execute paradigm, where future visual trajectories are first generated and then used for action prediction, or jointly model future video and actions within a shared generative process. Our work is most closely related to this line, but differs in focus: rather than proposing another imagine-then-execute WAM, we study whether the gains of WAMs come primarily from video co-training during training or from explicit future imagination during inference.

Our work is also related to recent efforts that exploit video modeling for action prediction while reducing or bypassing explicit test-time video synthesis. VPP~\cite{hu2024video} conditions robot policies on predictive visual representations extracted from a video diffusion model, while UVA~\cite{li2025unified} jointly models video and action and skips video decoding at test time for faster inference. Compared with these works, our focus is on disentangling the relative roles of training-time video co-training and test-time future imagination through controlled variants under a shared framework.

%% file: method.tex
\section{Method}
\subsection{Problem Formulation}

We consider embodied policy learning from visual observations and language instructions. Let $o$ denote the current observation, $l$ denote the task instruction, and $a_{1:H}$ denote an action chunk of horizon $H$. A standard visuomotor policy models the conditional distribution
\begin{equation}
    p(a_{1:H} \mid o, l),
\end{equation}
which directly maps the current perceptual context to a sequence of actions. World Action Models (WAMs) augment this formulation by introducing future visual observations as an intermediate variable. Let $v_{1:T}$ denote future visual observations over a prediction horizon $T$. Many existing WAMs follow an \emph{imagine-then-execute} factorization:
\begin{equation}
    p(a_{1:H} \mid o, l)
    =
    \int p(v_{1:T} \mid o, l)\,
    p(a_{1:H} \mid o, l, v_{1:T})\, dv_{1:T},
\end{equation}
where the model first predicts future observations and then conditions action generation on the imagined future. In practice, this is typically implemented either by jointly denoising future video and actions within a shared model, or by first generating future video and then feeding it to an inverse dynamics or action prediction module. Some prior WAMs further wrap this formulation in an outer auto-regressive rollout, which we omit here for simplicity and controlled comparison.

Our starting point is the observation that the effectiveness of WAMs may arise from two distinct factors: \emph{(i)} the video prediction objective used during training, which can encourage the model to learn physically meaningful latent representations, and \emph{(ii)} explicit future generation during inference, which may provide additional foresight for action prediction. Existing WAM formulations usually couple these two factors, since the same model both learns from future video prediction and explicitly synthesizes future observations at test time. We design Fast-WAM to decouple these two factors. During training, it retains world modeling as a co-training signal; during inference, however, it does not explicitly generate future observations. Instead, Fast-WAM predicts actions directly from the current observation and instruction,
\begin{equation}
    p_\theta(a_{1:H} \mid o, l),
\end{equation}
while using latent world representations shaped by video co-training. In this sense, Fast-WAM has a direct-policy interface at test time, similar to standard VLA policies, while its representation learning remains grounded in WAM-style video modeling during training. Formally, let $z(o, l)$ denote the latent world representation produced by the video backbone conditioned on the current context. Fast-WAM uses this representation to parameterize the action distribution,
\begin{equation}
    p_\theta(a_{1:H} \mid o, l)
    =
    p_\theta(a_{1:H} \mid z(o, l)).
\end{equation}
The key difference from imagine-then-execute WAMs is that $z(o, l)$ is obtained by a single forward encoding pass, rather than by explicitly sampling or denoising future observations $v_{1:T}$ at inference time.

\subsection{Model Architecture}

\paragraph{Overview.}
Fast-WAM is designed to preserve the training benefits of world modeling while removing the inference cost of explicit future imagination. During training, it jointly learns action prediction and video modeling, encouraging the visual backbone to capture physically meaningful motion and interaction structure. During inference, Fast-WAM does not explicitly generate future observations. Instead, it keeps only the clean latent tokens of the first observation frame, processes them with the video model in a single forward pass, and uses the resulting latent world representation for direct action generation. This gives Fast-WAM a direct-policy interface at test time while retaining WAM-style video supervision during training.

\begin{figure}[t]
    \centering
    \begin{subfigure}[t]{0.6\linewidth}
        \centering
        \includegraphics[width=\linewidth]{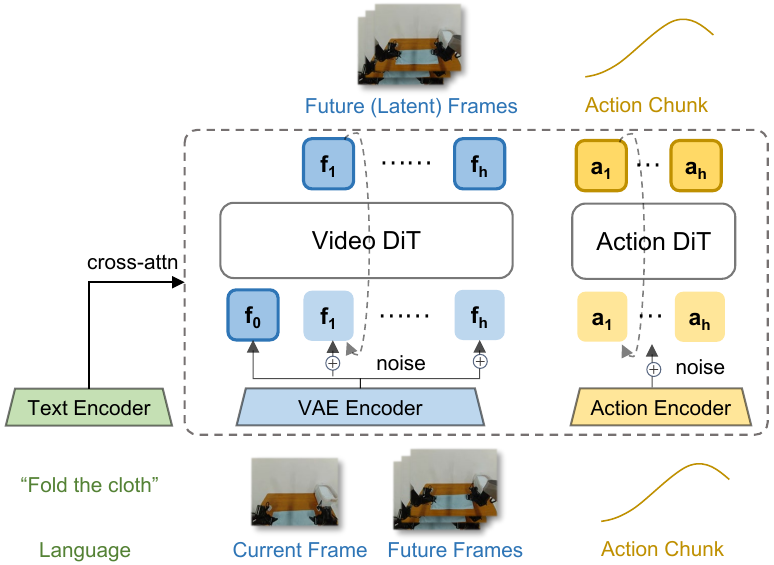}
        \caption{Fast-WAM model architecture.}
        \label{fig:model_arch}
    \end{subfigure}
    \hspace{0.05\linewidth}
    \begin{subfigure}[t]{0.2\linewidth}
        \centering
        \includegraphics[width=\linewidth]{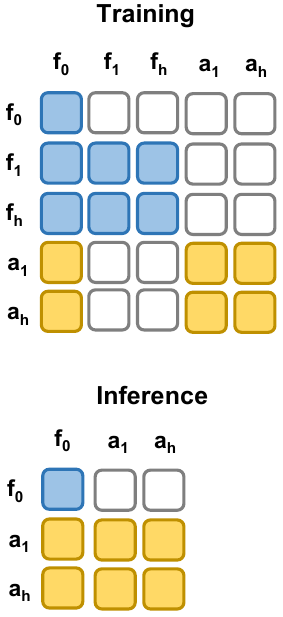}
        \caption{Training and inference masks.}
        \label{fig:mask}
    \end{subfigure}
    \caption{Fast-WAM architecture and the structured attention mask used to disentangle video co-training from action generation.}
    \label{fig:fastwa_arch}
\end{figure}

\paragraph{Architecture.}
Fast-WAM is built on top of the video Diffusion Transformer (DiT) from Wan2.2-5B~\cite{wan2025}, which serves as the world modeling backbone. We also reuse its pretrained text encoder and video VAE: task language is encoded by the built-in T5 encoder and provided to all tokens through cross-attention, while visual observations are mapped into latent video tokens by the pretrained VAE. On top of this backbone, we introduce an action expert DiT for action chunk generation. The full model is organized as a Mixture-of-Transformer (MoT) architecture with shared attention between the video and action branches, as illustrated in Figure~\ref{fig:model_arch}.

We organize the input tokens into three groups: clean latent tokens of the first observation frame, which serve as the shared visual anchor; noisy latent tokens of future video frames, which are used only during training for video modeling; and action tokens, which are processed by the action expert for action generation. All token groups attend to the language embeddings through cross-attention. A structured attention mask controls the information flow between these groups. During training, future noisy video tokens attend bidirectionally within the video branch and can access the clean first-frame tokens; action tokens attend bidirectionally within the action branch and can also access the clean first-frame tokens. Crucially, action tokens cannot attend to future video tokens, and the clean first-frame tokens do not attend to any other tokens. This ensures that both video modeling and action prediction are grounded in the same visual context while preventing future information from leaking into the action branch. We provide the full training and inference masks in Figure~\ref{fig:mask}.

At inference time, Fast-WAM removes the future video branch entirely: only the clean first-frame latent tokens are retained and passed through the video backbone once to produce latent world features for the action expert. Since no future noisy video tokens are instantiated and no explicit future video denoising is performed, Fast-WAM incurs substantially lower inference cost than standard imagine-then-execute WAMs.

\paragraph{Training objective.}
Fast-WAM is trained with a joint flow matching objective over action tokens and future video latents. Given a target variable $y$ (either an action chunk $a_{1:H}$ or future video latents $z_{1:T}$), we sample Gaussian noise $\epsilon \sim \mathcal{N}(0, I)$ and a time step $t \in(0,1)$, and construct the interpolated sample
\begin{equation}
    y_t = (1-t)\, y + t\, \epsilon.
\end{equation}
The model is trained to predict the corresponding velocity field with a standard flow matching objective
\begin{equation}
    \mathcal{L}_{\mathrm{FM}}(y)
    =
    \mathbb{E}_{y,\,\epsilon,\,t}
    \left[
    \left\|
    f_\theta(y_t,\, t,\, o,\, l)
    -
    (\epsilon - y)
    \right\|_2^2
    \right].
\end{equation}

We instantiate this objective for both action generation and video co-training. For action prediction, we set $y=a_{1:H}$ and optimize
\begin{equation}
    \mathcal{L}_{\mathrm{act}} = \mathcal{L}_{\mathrm{FM}}(a_{1:H}).
\end{equation}
For video co-training, we set $y=z_{1:T}$, where $z_{1:T}$ denotes the latent tokens of future video frames produced by the pretrained VAE, and optimize
\begin{equation}
    \mathcal{L}_{\mathrm{vid}} = \mathcal{L}_{\mathrm{FM}}(z_{1:T}).
\end{equation}
The overall training objective is
\begin{equation}
    \mathcal{L}
    =
    \mathcal{L}_{\mathrm{act}}
    +
    \lambda \mathcal{L}_{\mathrm{vid}},
\end{equation}
where $\lambda$ balances action learning and video co-training.

\subsection{Controlled Variants for Disentangled WAM Design}
\label{sec:controlled_variants}

To answer our central question, namely whether the benefit of WAMs comes primarily from video co-training during training or from explicit future imagination during inference, we design a set of controlled variants under a shared implementation framework. We instantiate representative imagine-then-execute design patterns from recent WAMs while keeping the backbone, tokenization, and training recipe as aligned as possible. This controlled setup allows us to isolate the contribution of test-time future generation from that of the video co-training objective itself.

As illustrated in Figure~\ref{fig:three_types}(A) and (B), we consider two representative imagine-then-execute variants that capture the dominant design patterns in recent WAMs. The first variant, named Fast-WAM-Joint, follows the joint-generation paradigm, where future video tokens and action tokens are denoised together within a shared model, so that action generation remains coupled with future video modeling throughout the denoising process~\cite{ye2026worldactionmodelszeroshot, zhu2025unifiedworldmodelscoupling, bi2025motusunifiedlatentaction}. The second variant, named Fast-WAM-IDM, follows the video-then-action paradigm, where future video tokens are generated first from the current observation and language context, and action prediction is then conditioned on the resulting future representation~\cite{lingbot-va2026, feng2025vidarembodiedvideodiffusion, du2023learninguniversalpoliciestextguided}. In both cases, we preserve the defining inference structure of the corresponding paradigm, together with key training choices used in recent WAMs, while implementing them within our shared framework for controlled comparison with Fast-WAM.

We further construct a Fast-WAM variant without video co-training. This variant keeps the architecture and inference procedure unchanged, and removes only the video modeling objective during training. It therefore serves as a direct control for the role of video co-training itself. Together, these controlled variants isolate the two factors that are typically entangled in prior WAMs: video co-training during training and explicit future imagination during inference.

%% file: exp.tex
\section{Experiment}
\subsection{Implementation Details}
We use pretrained Wan2.2-5B~\cite{wan2025} as the backbone, including its video DiT, text encoder, and video VAE. The action expert shares the same architecture as the video branch but uses a reduced hidden dimension of $d_a=1024$, resulting in a 1B action expert and a total model size of 6B parameters. We set the action horizon to $h=32$. Video frames are temporally downsampled by $4\times$, resulting in 9 video frames per chunk. Images from multiple cameras are concatenated into a single image before being fed into the VAE.

We use the same flow-matching formulation for both the video and action branches. Following~\cite{wan2025}, we adopt a logit-normal distribution over $t$ as the noise schedule during both training and inference. During inference, we use 10 denoising steps with classifier-free guidance (CFG) scale set to 1.0. We use AdamW with a learning rate of $1\times10^{-4}$, weight decay of 0.01, and a cosine annealing schedule for all training settings. Training is conducted in mixed precision with gradient clipping at 1.0. All latency numbers are measured on a single NVIDIA RTX 5090D V2 32GB GPU.

We build Fast-WAM-Joint by allowing attention between video and action tokens. For Fast-WAM-IDM, we follow~\cite{lingbot-va2026} and apply noise augmentation to the ground-truth video tokens with probability $p=0.5$.

\subsection{Experiment Setup}
We evaluate Fast-WAM and its variants on both simulation benchmarks, namely LIBERO~\cite{liu2023libero} and RoboTwin 2.0~\cite{chen2025robotwin}, as well as on a real-world manipulation task.

\paragraph{LIBERO.}
We follow the standard benchmarking protocol and train our models on the four LIBERO suites: LIBERO-Spatial, LIBERO-Object, LIBERO-Goal, and LIBERO-Long. Each suite contains 500 demonstrations spanning 10 tasks. All models are trained for 20k steps. We report success rates for each suite, evaluated over a total of 2000 trials across 40 tasks with different random seeds.

\paragraph{RoboTwin 2.0.}
RoboTwin 2.0 is a challenging bimanual manipulation benchmark featuring more than 50 tasks that require coordinated dual-arm control. We follow the multi-task training setup of~\cite{bi2025motusunifiedlatentaction, lingbot-va2026}, where models are trained on a mixture of 2,500 demonstrations collected in clean scenes and 25,000 demonstrations collected under heavy scene randomization, spanning over 50 tasks. All models are trained for 30k steps. We report average success rates over 100 trials per task under both clean and randomized settings.

\begin{figure*}[tbp]
        \centering
\includegraphics[width=1.0\textwidth]{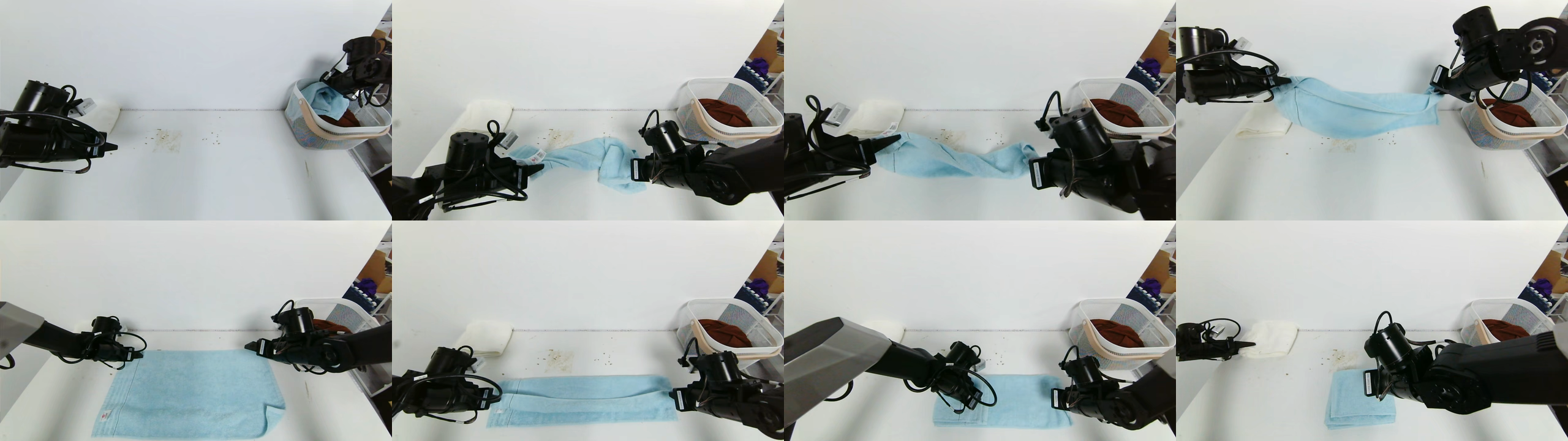} 
    \caption{Real-world towel-folding task on the Galaxea R1 Lite platform. Folding a deformable object requires long-horizon planning and precise closed-loop manipulation, making it a challenging benchmark for evaluating both task success and execution efficiency.}
    \label{fig:fold_bench}
\end{figure*}

\paragraph{Real-World Evaluation.}
We conduct real-world evaluation on towel folding, a long-horizon and challenging task that requires the policy to reason about the dynamics of deformable objects, as shown in Figure~\ref{fig:fold_bench}. We collect 60 hours of teleoperated demonstrations on the Galaxea R1 Lite platform. All models are trained for 30k steps. We report both the average success rate and the average completion time. The former measures whether the policy can eventually complete the folding task given sufficient time, while the latter reflects whether the policy learns an efficient execution strategy rather than relying on repeated trial-and-error corrections. For this task, completion time is therefore as important as success rate in assessing policy quality.

\subsection{Main Results}
\subsubsection{Overall comparison on simulation benchmarks}

\input{tables/tab_robotwin}
\input{tables/tab_libero}

Table~\ref{tab:robotwin} and Table~\ref{tab:libero} summarize the results on RoboTwin and LIBERO, respectively. Overall, Fast-WAM achieves performance comparable to state-of-the-art methods on both benchmarks without using embodied pretraining. Per-task results on RoboTwin are listed in Appendix Table~\ref{tab:robotwin_detail}.

On RoboTwin, Fast-WAM achieves 91.8\% success rate, exceeding all baselines without embodied pretraining and remaining highly competitive with the strongest pretrained WAMs. In particular, Fast-WAM substantially outperforms Motus~\cite{bi2025motusunifiedlatentaction} both with pretraining (87.8\%) and without pretraining (77.3\%), as well as LingBot-VA~\cite{lingbot-va2026} without pretraining (80.6\%), while remaining comparable to pretrained LingBot-VA (92.2\%). These results suggest that Fast-WAM can recover most of the benefit of prior WAM pipelines without relying on embodied pretraining.

On LIBERO, Fast-WAM also delivers strong overall performance, achieving an average success rate of 97.6\% without embodied pretraining. It outperforms the strong VLA baseline $\pi_{0.5}$ and remains competitive with pretrained WAM baselines such as LingBot-VA (98.5\%) and Motus (97.7\%). These results show that Fast-WAM is consistently effective across diverse simulation benchmarks, even without the embodied pretraining used by most prior WAM and VLA baselines.

\subsubsection{Controlled comparison with Fast-WAM variants}

We next compare Fast-WAM with the controlled variants introduced in Sec.~\ref{sec:controlled_variants} to disentangle the role of explicit future imagination at inference time from that of video co-training during training. Across both simulation benchmarks, the overall pattern is consistent: Fast-WAM remains comparable to the two imagine-then-execute variants, while removing video co-training leads to a much larger performance drop.

On RoboTwin, Fast-WAM achieves 91.8\% success rate, which is highly comparable to both Fast-WAM-Joint (90.6\%) and Fast-WAM-IDM (91.3\%). In contrast, removing video co-training reduces performance to 83.8\%, yielding a substantial gap relative to all three variants with video co-training. This result suggests that, on this benchmark, retaining the video modeling objective during training is far more important than explicitly generating future observations at test time. A similar trend holds on LIBERO. Fast-WAM reaches 97.6\% average success rate, remaining close to Fast-WAM-Joint (98.5\%) and Fast-WAM-IDM (98.0\%). By comparison, Fast-WAM without video co-training drops to 93.5\%, with especially visible degradation on the \textit{Spatial} and \textit{Long} subsets. Again, the gap caused by removing video co-training is clearly larger than the gap between Fast-WAM and the imagine-then-execute variants.

This pattern suggests that the main benefit of WAM-style training may lie less in whether, or how, future imagination is performed at test time, and more in the video prediction objective used to shape world-grounded representations during training.

\subsubsection{Real-world performance and efficiency}
\begin{figure*}[tbp]
        \centering
\includegraphics[width=0.95\textwidth]{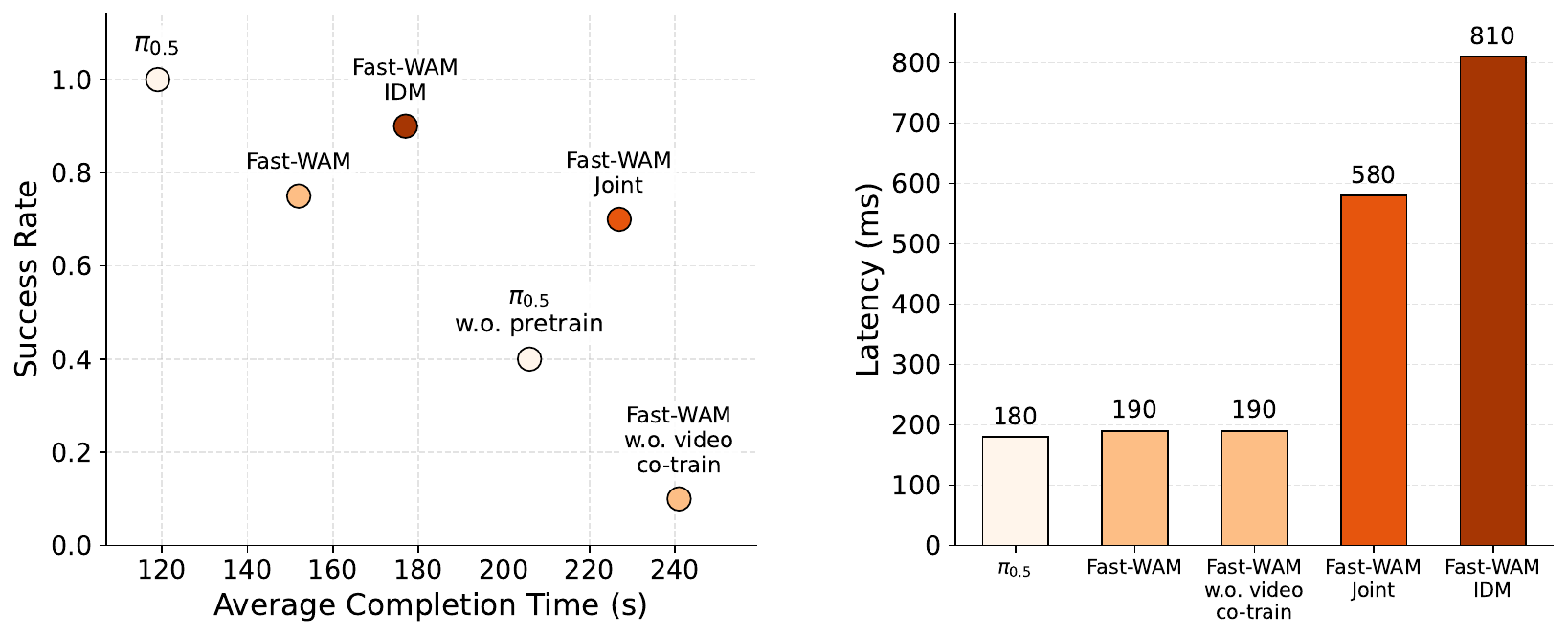} 
    \caption{Real-world results on the long-horizon towel-folding task. The left panel plots success rate against average completion time, where upper-left is better. The right panel compares inference latency. Fast-WAM achieves strong real-world performance with substantially lower latency than imagine-then-execute variants, while removing video co-training degrades both success rate and completion time.}
    \label{fig:realworld_latency}
\end{figure*}
We further evaluate Fast-WAM on a long-horizon real-world towel-folding task. We report both the average success rate and the average completion time, as they capture two equally important aspects of policy quality: whether the policy can eventually complete the task, and whether it does so efficiently rather than through repeated trial-and-error corrections. Results are shown in Figure~\ref{fig:realworld_latency}.

On this task, pretrained $\pi_{0.5}$ remains the strongest method, achieving the highest success rate together with the shortest completion time. Among the Fast-WAM family, the overall performance is comparable, with Fast-WAM-IDM achieving the highest success rate, while Fast-WAM attains a better completion time. Importantly, all Fast-WAM variants with video co-training substantially outperform $\pi_{0.5}$ without pretraining, indicating that WAM-style video co-training provides strong data efficiency even without embodied pretraining.

In contrast, removing video co-training causes a dramatic degradation on both metrics: Fast-WAM without video co-training drops to only 10\% success rate and also requires the longest completion time among all methods. This gap is much larger than the differences among the Fast-WAM variants themselves, suggesting that video co-training is the dominant factor behind strong real-world performance, whereas the effect of test-time future imagination is comparatively limited.

In terms of runtime, Fast-WAM retains low inference latency (190\,ms), whereas the imagine-then-execute variants are substantially slower, especially Fast-WAM-IDM at 810\,ms. This makes Fast-WAM a more favorable design point for real-world deployment, offering strong task performance together with much lower inference cost.

%% file: tables/tab_robotwin.tex
\begin{table}[t]
\centering
\small
\caption{Results on RoboTwin. Fast-WAM matches strong pretrained WAM baselines without using embodied pretraining, while the two imagine-then-execute variants remain highly comparable and removing video co-training causes a substantial drop.}
\label{tab:robotwin}
\resizebox{0.8\columnwidth}{!}{
\begin{tabular}{l|c|cc|c}
\toprule
Method & Embodied PT. & Clean & Rand. & \textbf{Average} \\
\midrule
% X-VLA & \cmark & 72.80 & 72.84 & 72.8 \\
$\pi_{0}$~\cite{black2024pi_0} & \cmark & 65.92 & 58.40 & 62.2 \\
$\pi_{0.5}$~\cite{intelligence2025pi_} & \cmark & 82.74 & 76.76 & 79.8 \\
Motus~\cite{bi2025motusunifiedlatentaction} & \cmark & 88.66 & 87.02 & 87.8 \\
Motus from WAN2.2 & \xmark & 77.56 & 77.00 & 77.3 \\
LingBot-VA~\cite{lingbot-va2026} & \cmark & 92.90 & 91.50  & \textbf{92.2} \\
LingBot-VA from WAN2.2 & \xmark & 80.60 & -- & 80.6 \\
\textbf{Fast-WAM (Ours)} & \xmark & 91.88 & 91.78 & \underline{91.8} \\
\midrule
\multicolumn{5}{l}{\textit{Fast-WAM Variants}} \\
\cmidrule(lr){1-5}
Fast-WAM & \xmark & 91.88 & 91.78 & \textbf{91.8} \\
Fast-WAM-Joint & \xmark & 90.84 & 90.32 & 90.6 \\
Fast-WAM-IDM & \xmark & 91.16 & 91.34 & \underline{91.3} \\
Fast-WAM w.o. video co-train & \xmark & 82.76 & 84.80 & 83.8 \\
\bottomrule
\end{tabular}
}
\end{table}

%% file: tables/tab_libero.tex
\begin{table*}[t]
\centering
\small
\caption{Results on LIBERO. Fast-WAM achieves competitive overall performance without embodied pretraining, remains close to both imagine-then-execute variants, and outperforms the no-video-co-training ablation by a clear margin.}
\label{tab:libero}
\resizebox{0.95\textwidth}{!}{
\begin{tabular}{l|c|cccc|c}
\toprule
Method & Embodied PT. & Spatial & Object & Goal & Long & \textbf{Average} \\
\midrule
OpenVLA~\cite{kim2024openvla} & \cmark & 84.7 & 88.4 & 79.2 & 53.7 & 76.5 \\
$\pi_0$~\cite{black2024pi_0} & \cmark & 96.8 & 98.8 & 95.8 & 85.2 & 94.1 \\
$\pi_{0.5}$~\cite{intelligence2025pi_} & \cmark & 98.8 & 98.2 & 98.0 & 92.4 & 96.9 \\
LingBot-VA~\cite{lingbot-va2026} & \cmark & 98.5 & 99.6 & 97.2 & 98.5 & \textbf{98.5} \\
Motus~\cite{bi2025motusunifiedlatentaction} & \cmark & 96.8 & 99.8 & 96.6 & 97.6 & \underline{97.7} \\
\textbf{Fast-WAM (Ours)} & \xmark & 98.2 & 100.0 & 97.0 & 95.2 & 97.6 \\
\midrule
\multicolumn{7}{l}{\textit{Fast-WAM Variants}} \\
\cmidrule(lr){1-7}
Fast-WAM & \xmark & 98.2 & 100.0 & 97.0 & 95.2 & 97.6 \\
Fast-WAM-Joint & \xmark & 99.6 & 99.4 & 98.2 & 96.8 & \textbf{98.5} \\
Fast-WAM-IDM & \xmark & 98.8 & 97.8 & 97.8 & 97.6 & \underline{98.0} \\
Fast-WAM w.o. video co-train & \xmark & 89.2 & 99.2 & 95.4 & 90.0 & 93.5 \\
\bottomrule
\end{tabular}
}
\end{table*}

%% file: conclusion.tex
\section{Conclusion}

In this paper, we revisited a basic question in World Action Models: whether their gains come primarily from explicit future imagination at test time or from video modeling during training. To study this question, we introduced Fast-WAM, a WAM architecture that retains video co-training during training while skipping future prediction at inference time, enabling direct action generation from world-grounded latent representations.
Across simulation benchmarks and real-world robotic tasks, Fast-WAM achieves strong performance without embodied pretraining while running in real time. More importantly, controlled comparisons show that Fast-WAM remains competitive with imagine-then-execute variants, whereas removing video co-training leads to a much larger degradation. These results suggest that the main value of video prediction in WAMs may lie more in learning better world representations during training than in generating future observations at test time.
An important direction for future work is to study the effect of larger-scale pretraining data and model scaling on this design.

%% file: tables/tab_robotwin_detail.tex
\begin{table*}[t]
\centering
\tiny
\caption{Per-task success rates on RoboTwin under clean and randomized evaluation settings. }
\label{tab:robotwin_detail}
\setlength{\tabcolsep}{2pt}
\resizebox{\textwidth}{!}{
\begin{tabular}{lcc|cc|cc|cc|cc|cc|cc}
\toprule
Task & \multicolumn{2}{c}{\textbf{Fast-WAM (Ours)}} & \multicolumn{2}{c}{Fast-WAM-Joint} & \multicolumn{2}{c}{Fast-WAM-IDM} & \multicolumn{2}{c}{\makecell{Fast-WAM\\w.o. co-train}} & \multicolumn{2}{c}{LingBot-VA} & \multicolumn{2}{c}{$\pi_{0.5}$} & \multicolumn{2}{c}{Motus} \\
\cmidrule(lr){2-3}\cmidrule(lr){4-5}\cmidrule(lr){6-7}\cmidrule(lr){8-9}\cmidrule(lr){10-11}\cmidrule(lr){12-13}\cmidrule(lr){14-15}
 & Clean & Rand. & Clean & Rand. & Clean & Rand. & Clean & Rand. & Clean & Rand. & Clean & Rand. & Clean & Rand. \\
\midrule
Adjust Bottle & \textbf{100} & \textbf{100} & 98 & 99 & 94 & 99 & 98 & \textbf{100} & 90 & 94 & \textbf{100} & 99 & 89 & 93 \\
Beat Block Hammer & 99 & 97 & \textbf{100} & \textbf{98} & 98 & \textbf{98} & 80 & 92 & 96 & \textbf{98} & 96 & 93 & 95 & 88 \\
Blocks Ranking RGB & \textbf{100} & \textbf{100} & \textbf{100} & \textbf{100} & \textbf{100} & 99 & 88 & 86 & 99 & 98 & 92 & 85 & 99 & 97 \\
Blocks Ranking Size & \textbf{94} & \textbf{98} & 83 & 91 & 79 & 90 & 56 & 62 & \textbf{94} & 96 & 49 & 26 & 75 & 63 \\
Click Alarmclock & \textbf{100} & \textbf{100} & \textbf{100} & \textbf{100} & 98 & \textbf{100} & \textbf{100} & 98 & 99 & \textbf{100} & 98 & 89 & \textbf{100} & \textbf{100} \\
Click Bell & \textbf{100} & \textbf{100} & \textbf{100} & 98 & \textbf{100} & 96 & \textbf{100} & \textbf{100} & \textbf{100} & \textbf{100} & 99 & 66 & \textbf{100} & \textbf{100} \\
Dump Bin Bigbin & \textbf{97} & 96 & 95 & 95 & 93 & \textbf{98} & 92 & 94 & 89 & 96 & 92 & 97 & 95 & 91 \\
Grab Roller & \textbf{100} & \textbf{100} & \textbf{100} & \textbf{100} & \textbf{100} & \textbf{100} & \textbf{100} & \textbf{100} & \textbf{100} & \textbf{100} & \textbf{100} & \textbf{100} & \textbf{100} & \textbf{100} \\
Handover Block & 95 & 81 & 93 & 91 & 97 & \textbf{94} & 58 & 46 & \textbf{99} & 78 & 66 & 57 & 86 & 73 \\
Handover Mic & 99 & \textbf{100} & \textbf{100} & \textbf{100} & 98 & 99 & \textbf{100} & \textbf{100} & 94 & 96 & 98 & 97 & 78 & 63 \\
Hanging Mug & 58 & \textbf{62} & \textbf{71} & 56 & 66 & \textbf{62} & 28 & 40 & 40 & 28 & 18 & 17 & 38 & 38 \\
Lift Pot & \textbf{100} & \textbf{100} & \textbf{100} & \textbf{100} & \textbf{100} & \textbf{100} & 92 & 90 & \textbf{100} & 99 & 96 & 85 & 96 & 99 \\
Move Can Pot & 90 & 88 & \textbf{97} & 99 & \textbf{97} & \textbf{100} & 80 & 68 & 94 & 97 & 51 & 55 & 34 & 74 \\
Move Pillbottle Pad & \textbf{100} & 99 & 99 & \textbf{100} & 98 & \textbf{100} & 88 & 96 & 99 & 99 & 84 & 61 & 93 & 96 \\
Move Playingcard Away & \textbf{100} & \textbf{100} & \textbf{100} & \textbf{100} & 99 & \textbf{100} & 94 & 96 & \textbf{100} & 99 & 96 & 84 & \textbf{100} & 96 \\
Move Stapler Pad & 77 & 64 & 85 & 81 & 89 & \textbf{85} & 64 & 78 & \textbf{91} & 79 & 56 & 42 & 83 & \textbf{85} \\
Open Laptop & 98 & \textbf{100} & 89 & 92 & 92 & 92 & \textbf{100} & 98 & 92 & 94 & 90 & 96 & 95 & 91 \\
Open Microwave & 62 & 45 & 3 & 14 & 54 & 53 & 46 & 52 & 82 & 86 & 34 & 77 & \textbf{95} & \textbf{91} \\
Pick Diverse Bottles & 80 & 85 & 86 & 87 & 87 & 89 & 58 & 62 & 89 & 82 & 81 & 71 & \textbf{90} & \textbf{91} \\
Pick Dual Bottles & \textbf{100} & 96 & 98 & \textbf{99} & \textbf{100} & 98 & 80 & 74 & \textbf{100} & \textbf{99} & 93 & 63 & 96 & 90 \\
Place A2B Left & 95 & 93 & 96 & \textbf{96} & \textbf{97} & \textbf{96} & 84 & 92 & \textbf{97} & 93 & 87 & 82 & 88 & 79 \\
Place A2B Right & 93 & \textbf{99} & 95 & 95 & 94 & 98 & 88 & 84 & \textbf{97} & 95 & 87 & 84 & 91 & 87 \\
Place Bread Basket & 91 & 93 & 89 & 94 & 91 & \textbf{97} & 74 & 76 & \textbf{97} & 95 & 77 & 64 & 91 & 94 \\
Place Bread Skillet & 90 & 93 & 90 & 93 & 90 & \textbf{95} & \textbf{98} & 84 & 95 & 90 & 85 & 66 & 86 & 83 \\
Place Burger Fries & 96 & 99 & \textbf{100} & \textbf{100} & 97 & 99 & 94 & 96 & 97 & 95 & 94 & 87 & 98 & 98 \\
Place Can Basket & 71 & 69 & 50 & 23 & 37 & 28 & 72 & 72 & \textbf{81} & \textbf{84} & 62 & 62 & \textbf{81} & 76 \\
Place Cans Plasticbox & 99 & 96 & 98 & 98 & 98 & 96 & 98 & 96 & \textbf{100} & \textbf{99} & 94 & 84 & 98 & 94 \\
Place Container Plate & 96 & \textbf{100} & 99 & 98 & \textbf{100} & 96 & 94 & 98 & 99 & 97 & 99 & 95 & 98 & 99 \\
Place Dual Shoes & \textbf{94} & 88 & 93 & \textbf{89} & 85 & 87 & 80 & 74 & \textbf{94} & \textbf{89} & 75 & 75 & 93 & 87 \\
Place Empty Cup & \textbf{100} & \textbf{100} & \textbf{100} & \textbf{100} & \textbf{100} & \textbf{100} & \textbf{100} & \textbf{100} & \textbf{100} & \textbf{100} & \textbf{100} & 99 & 99 & 98 \\
Place Fan & 96 & \textbf{96} & \textbf{99} & \textbf{96} & 97 & 95 & 80 & 88 & \textbf{99} & 93 & 87 & 85 & 91 & 87 \\
Place Mouse Pad & 83 & 89 & 96 & 91 & \textbf{97} & 93 & 64 & 76 & 93 & \textbf{96} & 60 & 39 & 66 & 68 \\
Place Object Basket & 89 & 88 & 86 & 81 & 87 & 82 & 82 & \textbf{90} & \textbf{91} & 88 & 80 & 76 & 81 & 87 \\
Place Object Scale & 90 & 97 & 96 & \textbf{99} & \textbf{99} & \textbf{99} & 86 & 80 & 96 & 95 & 86 & 80 & 88 & 85 \\
Place Object Stand & 90 & 94 & 92 & 98 & 96 & \textbf{100} & 82 & 92 & \textbf{99} & 96 & 91 & 85 & 98 & 97 \\
Place Phone Stand & 97 & 99 & \textbf{100} & \textbf{100} & 99 & 99 & 90 & 92 & 97 & 97 & 81 & 81 & 87 & 86 \\
Place Shoe & 96 & \textbf{99} & 95 & 97 & 95 & 98 & 92 & 90 & 98 & 98 & 92 & 93 & \textbf{99} & 97 \\
Press Stapler & 90 & 97 & 52 & 50 & 50 & 57 & 80 & 80 & 85 & 82 & 87 & 83 & \textbf{93} & \textbf{98} \\
Put Bottles Dustbin & 95 & 90 & 93 & \textbf{95} & \textbf{97} & 92 & 78 & 88 & 87 & 91 & 84 & 79 & 81 & 79 \\
Put Object Cabinet & 94 & 89 & \textbf{95} & \textbf{90} & 93 & \textbf{90} & 88 & 84 & 85 & 87 & 80 & 79 & 88 & 71 \\
Rotate QRcode & 93 & 89 & 91 & \textbf{92} & 91 & 86 & 82 & 78 & \textbf{96} & 91 & 89 & 87 & 89 & 73 \\
Scan Object & 89 & \textbf{92} & 92 & \textbf{92} & 93 & 90 & 78 & 86 & \textbf{96} & 91 & 72 & 65 & 67 & 66 \\
Shake Bottle & \textbf{100} & \textbf{100} & \textbf{100} & \textbf{100} & \textbf{100} & \textbf{100} & \textbf{100} & \textbf{100} & \textbf{100} & 97 & 99 & 97 & \textbf{100} & 97 \\
Shake Bottle Horizontally & \textbf{100} & \textbf{100} & \textbf{100} & \textbf{100} & \textbf{100} & \textbf{100} & \textbf{100} & \textbf{100} & \textbf{100} & 99 & 99 & 99 & \textbf{100} & 98 \\
Stack Blocks Three & 95 & 97 & 98 & 97 & \textbf{99} & 95 & 90 & 90 & \textbf{99} & \textbf{98} & 91 & 76 & 91 & 95 \\
Stack Blocks Two & \textbf{100} & \textbf{100} & \textbf{100} & \textbf{100} & \textbf{100} & \textbf{100} & \textbf{100} & 98 & \textbf{100} & 98 & 97 & \textbf{100} & \textbf{100} & 98 \\
Stack Bowls Three & 80 & 81 & 84 & 86 & 85 & 83 & 66 & 82 & \textbf{86} & 83 & 77 & 71 & 79 & \textbf{87} \\
Stack Bowls Two & 92 & \textbf{98} & 97 & 95 & 94 & 96 & 90 & \textbf{98} & 94 & \textbf{98} & 95 & 96 & \textbf{98} & \textbf{98} \\
Stamp Seal & 90 & 94 & 96 & \textbf{99} & \textbf{99} & 94 & 60 & 78 & 96 & 97 & 79 & 55 & 93 & 92 \\
Turn Switch & 61 & 59 & 73 & 72 & 59 & 74 & 66 & 66 & 44 & 45 & 62 & 54 & \textbf{84} & \textbf{78} \\
\midrule
\textbf{Average} & 91.88 & \textbf{91.78} & 90.84 & 90.32 & 91.16 & 91.34 & 82.76 & 84.80 & \textbf{92.90} & 91.50 & 82.74 & 76.76 & 88.66 & 87.02 \\
\bottomrule
\end{tabular}
}
\end{table*}